\documentclass[twocolumn]{svjour3}

\AtBeginDocument{%
	\providecommand\BibTeX{{%
			\normalfont B\kern-0.5em{\scshape i\kern-0.25em b}\kern-0.8em\TeX}}}

\usepackage{booktabs} % For formal tables
\usepackage{subfig}
\usepackage[ruled]{algorithm2e} % For algorithms
\usepackage{soul}
 \usepackage{amsmath}
 \usepackage{amssymb}
 \usepackage[ruled]{algorithm2e}
 
 \usepackage{epstopdf}
 \usepackage{wasysym}
 \usepackage{multirow}
 \usepackage{graphicx}
 \usepackage{caption}
 
 \graphicspath{{./fig/}}
 \usepackage{csquotes} 
 \usepackage[export]{adjustbox}
 \usepackage{subfig}

\newcommand{\Fscore}{\ensuremath{\textit{F-score}}}%
\newcommand{\etal}{\textit{et al.}}

\SetAlFnt{\small}
\SetAlCapFnt{\small}
\SetAlCapNameFnt{\small}
\SetAlCapHSkip{0pt}
\IncMargin{-\parindent}

\journalname{International Journal of Data Science and Analytics}
\begin{document}
% Title portion. Note the short title for running heads 

\title{A Fast Parallel Tensor Decomposition with Optimal Stochastic Gradient Descent: an Application in Structural Damage Identification  }

\author{Ali Anaissi *  \and Basem Suleiman   \and Seid Miad Zandavi }

 \institute{Ali Anaissi \at School of Computer Science, The University of Sydney. \\
 	\email{ali.anaissi@sydney.edu.au}    
 	\and Nguyen Lu Dang Khoa \at School of Computer Science, The University of Sydney.\\
 	\email{basem.suleiman@sydney.edu.au}
 	\and Yang Wang \at School of Computer Science, The University of Sydney.\\
 	\email{miad.zandavi@sydney.edu.au}	}
 
\maketitle

\begin{abstract}	
	Structural Health Monitoring (SHM) provides an economic approach
	which aims to enhance understanding the behavior of structures
	by continuously collects data through multiple networked
	sensors attached to the structure. This data is then utilized to gain
	insight into the health of a structure and make timely and economic
	decisions about its maintenance. The generated SHM sensing data
	is non-stationary and exists in a correlated multi-way form which
	makes the batch/off-line learning and standard two-way matrix
	analysis unable to capture all of these correlations and relationships.
	In this sense, the online tensor data analysis has become an essential
	tool for capturing underlying structures in higher-order datasets
	stored in a tensor $\mathcal{X}  \in  \mathbb{R} ^{I_1 \times \dots   \times I_N} $. The CANDECOMP/PARAFAC
	(CP) decomposition has been extensively studied and applied to
	approximate X by N loading matrices A(1), . . . ,A(N) where N represents
	the order of the tensor. 	We propose a novel algorithm, FP-CPD, to parallelize  the CANDECOMP/PARAFAC (CP)  decomposition  of a tensor  $\mathcal{X}  \in  \mathbb{R} ^{I_1 \times \dots   \times I_N} $.  Our approach is based  on stochastic gradient descent (SGD) algorithm  which   allows us to parallelize the learning process and it is very useful in  online setting since it  updates $\mathcal{X}^{t+1}$ in one single step. Our SGD algorithm is augmented with Nesterov's Accelerated Gradient (NAG) and perturbation methods to accelerate and guarantee convergence. The experimental results  using laboratory-based and real-life structural datasets indicate fast convergence and good scalability.

\end{abstract}




\section{Introduction}

There has been an exponential growth of data which is generated by the accelerated use of modern computing paradigms. A prominent example of such paradigms is the Internet of Things (IoTs) in which everything is envisioned to be connected to the Internet. One of the most promising technology transformations of IoT is a smart city. In such cities, enormous number of connected sensors and devices continuously collect massive amount of data about things such as city infrastructure to analyse and gain insights on how to manage the city efficiently in terms of resources and services. 

The adoption of smart city paradigm will result in massive increase of data volume (data collected from a large number of sensors) as well as a number of data features which increase data dimensionality. To make prices and in-depth insights from such data, advanced and efficient techniques including multi-way data analysis were recently adopted by research communities.  

The concept of multi-way data analysis was introduced by Tucker in 1964 as an extension of standard two-way data analysis to analyze multidimensional data known as tensor \cite{kolda2009tensor}.  It is often used when traditional two-way data analysis methods such as Non-negative Matrix Factorization (NMF), Principal Component Analysis (PCA) and Singular Value Decomposition (SVD) are not capable of capturing the underlying structures inherited in multi-way data~\cite{Cichocki2015}. In the realm of multi-way data, tensor decomposition methods such as $Tucker$ and $CANDECOMP/PARAFAC$ (CP) \cite{kolda2009tensor,Rendle2009} have been extensively studied and applied in various fields including signal processing~\cite{DeLathauwer1996}, civil engineer ~\cite{Khoa2017}, recommender systems  \cite{Rendle2009},  and  time series analysis~\cite{Cong2015}. The CP decomposition has gained much popularity for analyzing multi-way data due to its ease of interpretation. For example, given a tensor $\mathcal{X} \in \mathbb{R} ^{I_1 \times \dots \times I_N} $, CP method decomposes $\mathcal{X}$ by $N$ loading matrices $A^{(1)}, \dots, A^{(N)}$ each represents one mode explicitly, where $N$ is the tensor order and each matrix $A$ represents one mode explicitly. In contrast to  $Tucker$ method, the three modes can interact with each other making it difficult to interpret the resultant matrices. 

The CP decomposition approach often uses the Alternating Least Squares (ALS) method to find the solution for a given tensor. The ALS method follows the batch mode training process which iteratively solves each component matrix by fixing all the other components, then it repeats the procedure until it converges \cite{khoa2017smart}. However, ALS can lead to sensitive solutions  \cite{elden1980perturbation}\cite{anaissi2018regularized}. Moreover, in the domain of big data and IoTs such as smart cities, the ALS method raises many challenges in dealing with data that is continuously measured at high velocity from different sources/locations and dynamically changing over time. For instance,  a structural health monitoring (SHM) data can be represented in a three-way form as  $location \times feature \times time$ which represents a large number of vibration responses measured over time by many sensors attached to a structure at different locations. This type of data can be found in many other application domains including  \cite{acar2009unsupervised,sun2008incremental,kolda2008scalable,anaissi2018tensor}. The iterative nature of employed CP decomposition methods involve intensive computational processing in each iteration. A significant challenge arises in such algorithms (including ALS and its variations) when  the input tensor is sparse and has N dimension. This means as the dimensionality of the tensor increases, the calculations involved in the algorithm becomes computationally more expensive and thus incremental, parallel and distributed algorithms for CP decomposition becomes essential to achieving a  more reasonable performance This is especially the case in large applications and computing paradigms such as smart cites. 

The efficient processing of CP decomposition problem has been investigated with different hardware architecture and techniques including MapReduce structure \cite{Kang2012} and shared and distributed memory structures~\cite{Smith2015,Kaya2015}. Such approaches present algorithms that require alternating hardware architectures to enable parallel and fast execution of CP decomposition methods. The MapReduce and distributed computing approaches could also incur additional performance from network data communication and transfer. Our goal is to devise a parallel and efficient CP decomposition execution method with minimal hardware changes to the operating environment and without incurring additional performance resulting from new hardware architectures. %A naive approach to achieve this would be to recompute the CP decomposition from scratch for each new incoming  $X^{(t+1)}$. Therefore, this would become impractical and computationally expensive as new incoming datum would have a minimal effect on the current tensor.
%Zhou et al. \cite{zhou2016accelerating} proposed a method called onlineCP to address the problem of online CP decomposition using ALS algorithm. The method was able to incrementally update the temporal mode in multi-way data but failed for non-temporal modes \cite{khoa2017smart}. In recent years, several studies have been proposed to solve the CP decomposition using stochastic gradient descent (SGD) algorithm which has the capability to deal with big data and online learning. However, such methods are inefficient and impractical due to slow convergence, numerical uncertainty and non-convergence \cite{ge2015escaping,anandkumar2016efficient,maehara2016expected,rendle2010pairwise}. 
Thus, to address the aforementioned problems, we propose an efficient solver method called FP-CPD (Fast Parallel-CP Decomposition) for analyzing  large-scale high-order data in parallel based on stochastic gradient descent. The scope of this paper is smart cities and, in particular, SHM of infrastructure such as bridges. The novelty of our proposed method is summarized in the following contributions:
\begin{enumerate}
	
	\item \textbf{Parallel CP Decomposition.} Our FP-CPD method  is capable of efficiently learning large scale tensors in parallel and updating $\mathcal{X}^{(t+1)}$ in one step. 
	
	\item \textbf{Global convergence guarantee.} We followed the perturbation approach which adds a little noise to the gradient update step to reinforce the next update step to start moving away from a saddle point toward the correct direction. 
	
	\item \textbf{Optimal convergence rate.} Our method employs Nesterov's Accelerated Gradient (NAG) method into the SGD algorithm to optimally accelerate the convergence rate \cite{sutskever2013importance}.  It achieves a global convergence rate of $O(\frac{1}{T^2})$ comparing to  $O(\frac{1}{T})$ for traditional SGD.
	
	\item \textbf{Empirical analysis on structural  datasets.} We conduct experimental analysis using laboratory-based and real-life  datasets in the  field of SHM. The experimental analysis shows that our method can achieve more stable and fast tensor decomposition compared to other known existing online and offline methods.      
\end{enumerate}

The remainder of this paper is organized as follows.  Section \ref{s:related} introduces background knowledge and review of the related work. Section \ref{s:method} describes our novel FP-CPD algorithm for parallel CP decomposition based on SGD algorithm augmented with the NAG method and perturbation approach. Section \ref{s:motiv} presents the motivation of this work. Section \ref{s:results} shows the performance of D-CPD on structural datasets and presents our experimental results on both laboratory-based and real-life  datasets. The conclusion and discussion of future research work are presented in Section \ref{s:conclusion}.

\section{Background and Related work}
  \label{s:related}

\subsection{CP Decomposition}
Given a three-way tensor $\mathcal{X} \in \Re^{I \times J \times K} $,  CP  decomposes $\mathcal{X}$ into three  matrices  $A \in \Re^{I \times R}$, $B \in \Re^{J \times R} $and $ C \in \Re^{K \times R}$, where $R$ is the latent factors. It can be written as follows:

\begin{eqnarray}\label{eq:decomp}
\mathcal{X}_{(ijk)}  \approx \sum_{r=1}^{R}A_{ir} \circ B_{jr} \circ C_{kr}
\end{eqnarray}
where  "$\circ$" is a vector outer product.  $R$ is the latent element, $A_{ir}, B_{jr} $ and $C_{kr}$ are r-th columns of component 
matrices $A \in \Re^{I \times R}$, $B \in \Re^{J \times R} $and $ C \in \Re^{K \times R}$.
The main goal of  CP decomposition is   to decrease the sum  square error  between  the model and a given tensor $\mathcal{X}$. Equation \ref{eq:als} shows our loss function $L$ needs to be optimized.
 
\begin{eqnarray}\label{eq:als}	
L (\mathcal{X}, A, B, C) = \min_{A,B,C} \|  \mathcal{X} - \sum_{r=1}^R  \ A_{ir} \circ B_{jr} \circ C_{kr} \|^2_f,
\end{eqnarray}
where $\|\mathcal{X}\|^2_f$  is the  sum squares  of $\mathcal{X}$  and  the subscript $f$ is  the Frobenius norm. The loss function $L$ presented in Equation \ref{eq:als} is a non-convex problem with many local minima since it aims to optimize the sum squares of  three  matrices.  Several algorithms have been proposed to solve CP  decomposition \cite{symeonidis2008tag,lebedev2014speeding,rendle2009learning}. Among these algorithms, ALS  has been heavily employed which repeatedly solves each component matrix  by locking all other components until it   converges \cite{papalexakis2017tensors}. The rational idea of the least square algorithm is to set the partial  derivative of the loss function to zero with respect  to the parameter we need to minimize. Algorithm \ref{ALS} presents the detailed steps of ALS.

\begin{algorithm}
	%\renewcommand\thetable{Algorithm}
	%	\centering
	%
	%\label{ALS}
	\textbf{Alternating Least Squares\\}
	\textbf{Input}: Tensor $\mathcal{X} \in \Re^{I \times J \times K}  $, number of components $R$\\
	\textbf{Output}: Matrices  $A \in \Re^{I \times R}$, $B \in \Re^{J \times R} $ and  $ C \in \Re^{K \times R}$
	\begin{enumerate}
		\item[1:] Initialize $A,B,C$
		\item[2:] Repeat
		%	\begin{enumerate}
		{\setlength\itemindent{6pt}
			\item[3:] $A = \underset{A}{\arg\min} \frac{1}{2} \| X_{(1)} - A ( C \odot B)^T\|^2 $
			\item[4:] $B = \underset{B}{\arg\min} \frac{1}{2} \| X_{(2)} - B ( C \odot A)^T\|^2 $
			\item[5:] $C = \underset{C}{\arg\min} \frac{1}{2} \| X_{(3)} - C ( B \odot A)^T\|^2 $
			\item[]($X_{(i)} $ is the  unfolded matrix of $X$ in a current mode)	
		}
		%	\end{enumerate}
		\item[6:] until convergence
	\end{enumerate}
	\caption{  Alternating Least Squares for CP}
 	\label{ALS}
\end{algorithm}

Zhou \etal \cite{zhou2008large} suggest that ALS can be easily parallelized for matrix factorization methods, but its not scalable for large scale data especially when it  deals with multi-way tensor data.
Later Zhou \etal. \cite{zhou2016accelerating} proposed another method called onlineCP to address the problem of online  CP decomposition using ALS algorithm. The method was able to incrementally update the temporal mode in multi-way data but failed for non-temporal modes \cite{khoa2017smart} and not parallelized. 

%In recent years, several studies have been proposed to solve the CP decomposition using stochastic gradient descent (SGD) algorithm which has the capability to deal with big data and online learning. However, such methods are inefficient and impractical due to slow convergence, numerical uncertainty and non-convergence \cite{ge2015escaping,anandkumar2016efficient,maehara2016expected,rendle2010pairwise}. 

\subsection{Stochastic Gradient Descent}
A stochastic gradient descent algorithm is a key tool for optimization problems. Here, the aim is to optimize a loss function  $L(x,w)$, where $x$ is a data point  drawn from a  distribution $\mathcal{D}$ and $w$ is a variable. The stochastic optimization problem can be defined as follows:
\begin{eqnarray}\label{eq:sgd}
w =  \underset{w}{argmin} \;    \mathbb{E}[L(x,w)]
\end{eqnarray}
The stochastic gradient descent method solves the above problem defined in Equation \ref{eq:sgd} by  repeatedly updates $w$ to minimize $L(x,w)$. It starts with some initial value of $w^{(t)}$ and then repeatedly performs the update as follows:
\begin{eqnarray}\label{eq:sgdu}
w^{(t+1)} :=    w^{(t)}   + \eta   \frac{\partial L}{\partial w } (x^{(t)} ,w^{(t)} )
\end{eqnarray}
where $\eta$ is the learning rate and $x^{(t)}$ is a random sample drawn from the given distribution $\mathcal{D}$.
This method guarantees the convergence of the loss function $L$ to the global minimum when it is convex. However, it can be susceptible to many local minima and saddle points when the loss function exists in a non-convex setting. Thus it becomes an NP-hard problem. Note, the main bottleneck here is due to the existence of many saddle points and not to the local minima \cite{ge2015escaping}. This is because the rational idea of gradient algorithm depends only on the gradient information which may have $\frac{\partial L}{\partial u } = 0$ even though it is not at a minimum. 

Previous studies have used SGD for parallel matrix factorization. Gemulla \cite{gemulla2011large} proposed a new parallel method for matrix factorization using SGD. The authors indicate  the  method was able to  handle large scale data with fast convergence efficiently. Similarly, Chin \etal \cite{chin2015fast} proposed a fast parallel SGD method for matrix factorization in recommender systems. The method also applies  SGD in shared memory systems but with a careful consideration to the load balance of threads. Naiyang \etal \cite{guan2012nenmf}  applies  Nesterov's optimal gradient method to SGD for non-negative matrix factorization. This method accelerates the NMF process with less computational time. Similarly, Shuxin \etal \cite{zheng2017asynchronous} used an SGD algorithm for matrix factorization using Taylor expansion and Hessian information. They proposed a new asynchronous SGD algorithm to compensate for the delay resultant from a Hessian computation. 

Recently, SGD has attracted several researchers working on tensor decomposition.   For instance,  Ge \etal \cite{ge2015escaping} proposed a perturbed SGD (PSGD) algorithm for orthogonal tensor optimization. They presented several theoretical analysis that ensures convergence; however, the method is not applicable to non-orthogonal tensor. They also did not address the problem of slow convergence. Similarly, Maehara \etal  \cite{maehara2016expected} propose a new algorithm for CP decomposition based on a combination of SGD and ALS methods (SALS). The authors claimed the algorithm works well in terms of accuracy. Nevertheless,  its theoretical properties have not been completely proven and the saddle point problem was not addressed. Rendle and Thieme \cite{rendle2010pairwise} propose a pairwise interaction tensor factorization method based on  Bayesian personalized rank. The algorithm was designed to work only on three-way tensor data. To the best of our knowledge, this is the first work applies a parallel SGD algorithm augmented with Nesterov's optimal gradient and perturbation methods for fast parallel CP decomposition of multi-way tensor data.

\section{Fast Parallel CP Decomposition (FP-CPD)}
\label{s:method}

Given an $N^{th}$-order tensor $\mathcal{X} \in \mathbb{R}^{I_1 \times \dots \times I_N}$, we solve the CP decomposition by splitting the problem into a convex $N$ sub-problems since its loss function $L$ defined in Equation  \ref{eq:decomp} is non-convex problem which may have many local minima. In case of distributing  this solution, another challenge is raised where the value of the  $w^{(t)}$ must be globally updated before computing  $w^{(t+1)}$ where $w$ represents  $A, B$ and $C$. However, the structure and the process of tensor decomposition allows us to exploit this challenge. For illustration purposes, we present our FP-CPD method based on three-way tensor data. The same logic can be naturally extended to handle a higher-order tensor, though.

\vspace{1em} 
\begin{definition}
	\label{def1}
	Two training points $x_1 = (i_1,j_1,k_1) \in  \mathcal{X}$ and   $x_2 = (i_2,j_2,k_2) \in  \mathcal{X}$ are interchangeable with respect to the loss function $L$ defined in Equation  \ref{eq:decomp} if they are not sharing any dimensions, i.e, $i_1\neq i_2, j_1 \neq j_2$ and $k_1 \neq k_2$.
\end{definition}
\vspace{1em}

\begin{figure*}[!t]
	\centering
	\includegraphics[scale=0.8]{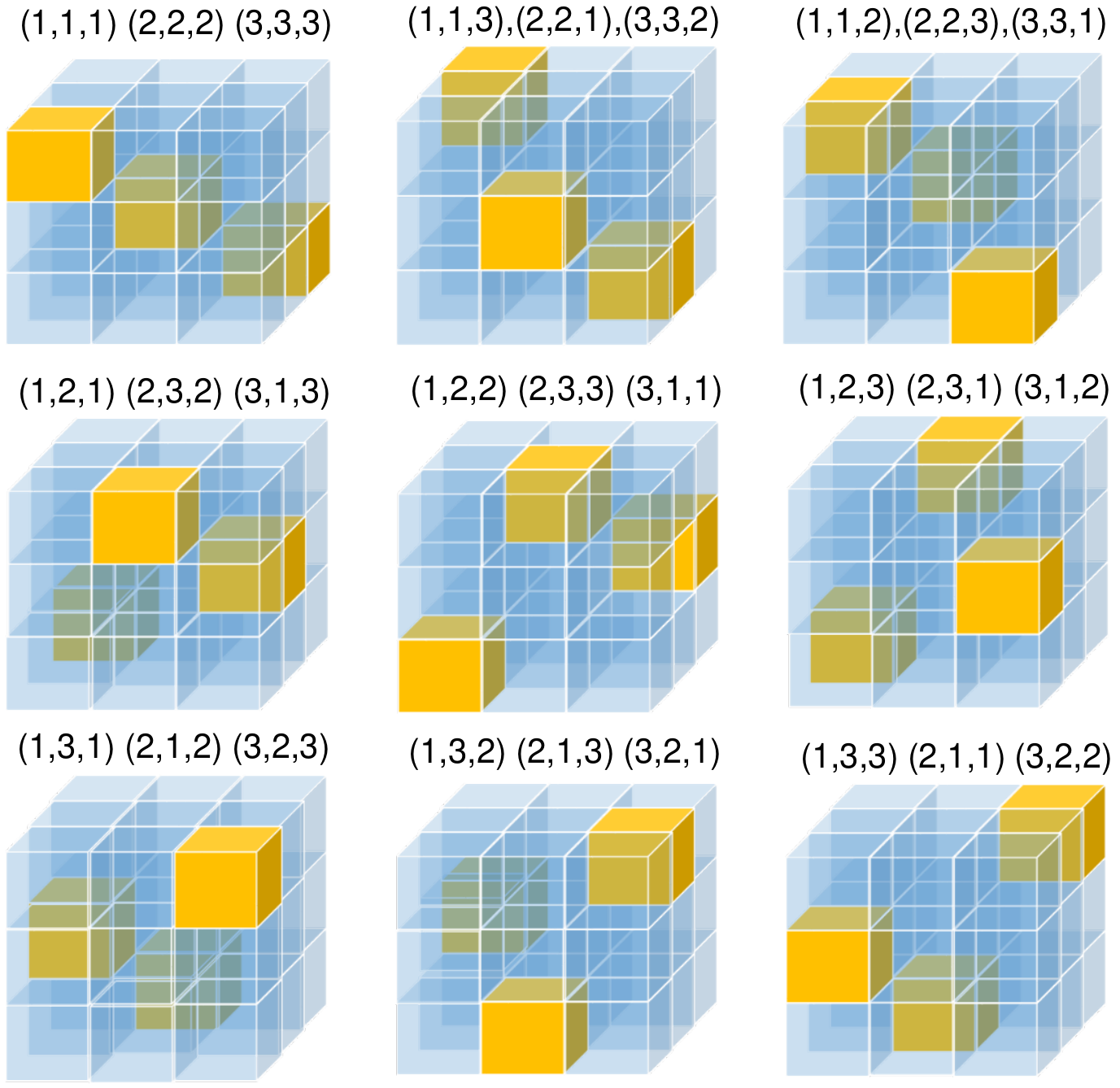}
	\caption{ Independent blocks for  $\mathcal{X} \in \Re^{3 \times 3 \times 3} $   }
	\label{tensor_para}
\end{figure*}

Based on Definition \ref{def1}, we develop a new algorithm, called FP-CPD, to carry the  tensor decomposition process in parallel.  The core idea of FP-CPD algorithm is to find and run the CPD in parallel by considering all the defined interchangeable training points in one single step without affecting the final outcome of $w$. Our FP-CPD algorithm partitions the training tensor $\mathcal{X} \in \Re^{I \times J \times K} $ into set of potentially independent blocks $\mathcal{X}_1,\dots, \mathcal{X}_b$.
Each block consists of $t$ interchangeable training points  which  are identified  by  finding all the the possible combinations of each dimension of a given tensor $\mathcal{X}$. To illustrate this process, we consider a three-order tensor $\mathcal{X} \in \mathbb{R}^{3 \times 3 \times 3}$ as shown in Figure \ref{tensor_para}. This tensor is partitioned into $d$ independent blocks which cover the entire given training data $\mathcal{D}_{b=1}^{d} \mathcal{X}_b$. The value of $d = \frac{i \times j \times k}{\min(i,j,k)}$. Each  $\mathcal{X}_b$ contains a parallelism parameter $p$ which deduces the possible number of  tasks that can be run in parallel. In our three-way tensor example $p =3$ interchangeable training points. 

\subsubsection{\textbf{The FP-CPD Algorithm }}

Given the set of independent blocks  $\mathcal{D}_{b=1}^{d} \mathcal{X}_b$, we can  decompose $\mathcal{X} \in \Re^{I \times J \times K} $ in parallel  into three matrices $A \in \Re^{I \times R}$, $B \in \Re^{J \times R} $ and $ C \in \Re^{K \times R}$, where $R$ is the latent factors. In this context, we reconstitute  our loss function defined in Equation \ref{eq:als} to be the sum of losses per block:$ L (A, B, C) = \sum_{b=1}^{d} L_b ( A, B, C) $. This new loss function provides  the rational of our parallel  CP decomposition which will allow SGD algorithm to learn all the  possible interchangeable data points within each block in parallel. Therefore, SGD computes the partial  derivative of the loss function $L_b (A, B, C) = \sum_{(i,j,k) \in \mathcal{D}_{b} } L_{i,j,k}(A, B, C)$ with respect to the three modes $A, B$ and $C$ alternatively as follows:
\begin{eqnarray}\label{eq:partial}
\frac{\partial L_b}{\partial A }(X^{(1)}; A)  =   (X^{(1)} -   A \times  (C \circ B)) \times (C \circ B) \nonumber\\
\frac{\partial L_b}{\partial B }(X^{(2)}; B)  =   (X^{(2)} -   B  \times (C \circ A)) \times (C \circ A)\\
\frac{\partial L_b}{\partial C }(X^{(3)}; C)  =   (X^{(3)} -   C \times  (B \circ A)) \times (B \circ A)\nonumber
\end{eqnarray}
where $X^{(i)}$ is an unfolding matrix of tensor $\mathcal{X}$ in mode $i$. The gradient update step for $A, B$ and $C$ are as follows:
\begin{eqnarray}\label{eq:update}
A^{(t+1)} :=    A^{(t)}   + \eta^{(t)}   \frac{\partial L_b}{\partial A } (X^{(1, t)} ;A^{(t)} )  \nonumber\\
B^{(t+1)} :=    B^{(t)}   + \eta^{(t)}   \frac{\partial L_b}{\partial B } (X^{(2, t)} ;B^{(t)} )  \\
C^{(t+1)} :=    C^{(t)}   + \eta^{(t)}   \frac{\partial L_b}{\partial C } (X^{(3, t)} ;C^{(t)} )  \nonumber
\end{eqnarray}

\subsubsection{\textbf{Convergence}} Regardless if we are applying parallel SGD or just SGD,  the partial derivative of SGD in non-convex setting may encounter data points with  $\frac{\partial L}{\partial w } = 0$ even though it is not at a global minimum. These data points are known as \textit{saddle points}  which may detente the optimization process to reach the desired local  minimum if not escaped  \cite{ge2015escaping}. These saddle points  can be  identified by studying the second-order derivative  (aka Hessian)  $\frac{\partial L}{\partial w }^2$. Theoretically, when the $\frac{\partial L}{\partial w }^2(x;w)\succ  0$, $x$ must be a local minimum; if $\frac{\partial L}{\partial w }^2(x;w) \prec 0$, then we are at a local maximum; if $\frac{\partial L}{\partial w }^2(x;w)$ has both positive and negative eigenvalues, the point is a saddle point. The second order methods guarantee  convergence, but the  computing of Hessian matrix $H^{(t)}$ is  high,  which makes the method infeasible for high dimensional data and online learning. Ge \etal \cite{ge2015escaping} show that saddle points are very unstable and can be escaped if we slightly perturb them with some noise. Based on this, we use the perturbation approach which adds Gaussian noise to the gradient. This reinforces the next update step to start moving away from that saddle point toward the correct direction. After a random perturbation, it is highly unlikely that the point remains in the same band and hence it can be efficiently escaped (i.e., no longer a saddle point). We further incorporate Nesterov's method into the perturbed-SGD algorithm to accelerate the convergence rate. Recently, Nesterov's Accelerated Gradient (NAG) \cite{nesterov2013introductory} has received much attention for solving convex optimization problems \cite{guan2012nenmf,nitanda2014stochastic,ghadimi2016accelerated}. It introduces a smart variation of momentum that works slightly better than standard momentum. This technique modifies the traditional SGD by introducing velocity $\nu$ and friction $\gamma$, which tries to control the velocity and prevents overshooting the valley while allowing faster descent. Our idea behind Nesterov's is to calculate the gradient at a position that we know our momentum is about to take us instead of calculating the gradient at the current position. In practice, it performs a simple step of gradient descent to go from $w^{(t)} $ to $w^{(t+1)}$, and then it shifts slightly further than $w^{(t+1)}$ in the direction given by $\nu^{(t-1)}$. In this setting, we model the gradient update step with NAG as follows: 
\begin{eqnarray}\label{eq:nagNe}
A^{(t+1)} :=    A^{(t)}   + \eta^{(t)}   \nu^{(A, t)} +  \epsilon - \beta ||A||_{L_{1,b}} 
\end{eqnarray}
where
\begin{eqnarray}\label{eq:velNe}
\nu^{(A, t)} :=    \gamma \nu^{(A, t-1)}  + (1-\gamma)  \frac{\partial L_b}{\partial A } (X^{(1, t)} ,A^{(t)} ) 
\end{eqnarray}
where $\epsilon$ is a Gaussian noise, $\eta^{(t)}$ is the step size,  and $||A||_{L_{1,b}}$ is the regularization and penalization parameter into the $L_1$ norms to achieve smooth representations of the outcome and thus bypassing the perturbation surrounding the local minimum problem. The updates for $(B^{(t+1)} , \nu^{(B, t)})$ and $(C^{(t+1)}  ,\nu^{(C, t)} )$ are similar to the aforementioned ones.
With NAG, our method achieves a global convergence rate of $O(\frac{1}{T^2})$ comparing to $O(\frac{1}{T})$  for traditional gradient descent. Based on the above models, we present our FP-CPD algorithm \ref{FP-CPD}. 

\begin{algorithm}
	%\renewcommand\thetable{Algorithm}
	%	\centering
	\caption{  FP-CPD algorithm}
	\label{FP-CPD}
	
	\textbf{Input}: Tensor $X \in \Re^{I \times J \times K}  $ , number of components $R$\\
	\textbf{Output}: Matrices  $A \in \Re^{I \times R}$, $B \in \Re^{J \times R} $ and  $ C \in \Re^{K \times R}$
	\begin{itemize}
		\item Initialize $A,B,C$
		\item Repeat
		%	\begin{enumerate}
		{
			\setlength\itemindent{10pt}{
				\item\textbf{Form} $d$ blocks $\{\mathcal{X}_1,\dots, \mathcal{X}_b\}$
				\item\textbf{for} $b=1, \dots, d$ \textbf{do}
				
				\setlength\itemindent{20pt}{
					
					\item $IP$ = Find all interchangeable data points in block \item[]$\mathcal{X}_b$ (Definition \ref{def1})
					
					\item \textbf{for each} $p$ in  $IP$ \textbf{do} {in parallel}			
					\setlength\itemindent{30pt}{
						
						%	\item Select training points in  $\mathcal{X}_b$
						\item Compute the partial derivative of $A, B$ and $C$ \item[]using Equation \ref{eq:partial}
						\item Compute $\nu$ of $A, B$ and $C$ using Equation \ref{eq:velNe}
						\item Update $A, B$ and $C$ using Equation \ref{eq:nagNe}			   
					}		
				}
				\item	\textbf{end for each}
				
			}

			\item \textbf{end for}
			%	\end{enumerate}
		}	
		\item \textbf{until convergence}
	\end{itemize}
	
\end{algorithm}

\section{Motivation}
  \label{s:motiv}
Numerous types of data are naturally structured as multi-way data. For instance, structural health monitoring (SHM) data can be represented in a three-way form as  $location \times feature \times time$. Arranging and analyzing the SHM data in a multidimensional form would allow us to  capture the correlation between sensors  at different locations and  at the same time which was not possible using the standard two-way matrix $time\times feature$. Furthermore, in SHM only positive data instances i.e healthy state are available. Thus, the problem becomes an anomaly detection problem in higher-order datasets.  Rytter \cite{rytter1993vibrational} affirms that damage identification also requires also damage localization and severity assessment which are considered much more complex than damage detection since they require a supervised learning approach \cite{worden2006application}. 

%To address the above problems in SHM applications, we employ our NeCPD method to learn from SHM data in multiple modes at the same time, and we use one-class SVM \cite{scholkopf2000support} as an anomaly detection method. The rationale behind one-class SVM is to map a positive data into a feature space using a kernel method. Recently, the Gaussian kernel has gained much popularity in many application domains. It has a parameter denoted $\sigma$ which may profoundly affect the performance of a one-class SVM by over-fitting or under-fitting the model. In our NeCPD approach, we use Edged Support Vector (ESV) algorithm \cite{anaissi2018gaussian} to tune $\sigma$ as it has the capability to work in a one-class learning setting.

Given a positive three-way SHM data $\mathcal{X} \in  \mathbb{R}^{feature \times location \times time}$, FP-CPD decomposes $\mathcal{X}$ into three matrices $A, B$ and $C$. The $C$ matrix represents the temporal mode where each row contains information about the vibration responses related to an event at time $t$. The analysis of this component matrix can help to detect the damage of the monitored structure. Therefore, we use the $C$ matrix to build a one-class anomaly detection model using only the positive  training events. For each new incoming $\mathcal{X}_{new}$, we update the three matrices $A, B$ and $C$ incrementally as described in Algorithm \ref{FP-CPD}. Then the constructed model estimates the agreement between the new event $C_{new}$ and the trained data. 

For damage localization,  we analyze the data in the location matrix $B$, where each row captures meaningful information for each sensor location. When the matrix $B$ is updated due to the arrival of a new event $\mathcal{X}_{new}$, we study the variation of the values in each row of matrix $B$ by computing the average distance from $B$'s row to $k$-nearest neighboring locations as an anomaly score for damage localization.
For severity assessment in damage identification, we study the decision values returned from the one-class  model. This is because a structure with more severe damage will behave much differently from a normal one.
\section{Evaluation}
\label{s:results}
In this section we presents the details of the experimental settings and the comparative analysis between our proposed FP-CPD algorithm and the alike parallel tensor decomposition algorithms; PSGD and SALS. We first analyze the  effectiveness and speed of the training process of the three algorithms based on four real-world datasets from SHM. We, then, evaluate the performance of our approach, along with other baselines using  the SHM datastes, in terms of damage detection, assessment and localization.  

\subsection{Experiment Setup and Datasets}
\label{sec:setup}
We conducted all our experiments  using a dual Intel Xeon processors with  32 GB  memory and 12 physical cores. We use R development environment to implement our FP-CPD algorithm and  PSGD and SALS algorithms with the help of the two packages  \textbf{rTensor} and \textbf{e1071} for tensor tools and one-class model.

We run our experiments on four real-world datasets, all of which inherently entails multi-way data structure. The datasets are collected from sensors that measure the health of building, bridge or road structures. Specifically, these datasets comprise of: 

\begin{enumerate}
	
	\item bridge structure measurement data collected from sensors attached to a cable-stayed bridge in  Western Sydney, Australia (BRIDGE) \cite{anaissi2018tensor}.  
	\item building structure measurement data collected from sensors attached to a specimen building structure obtained from Los Alamos National Laboratory (LANL) \cite{larson1987alamos} (BUILDING).
	\item measurements data collected from loop detectors in Victoria, Australia (ROAD) \cite{schimbinschi2015traffic}.  
	\item road measurements collected from sensors attached to two buses travelling through routes in the southern region of New South Wales, Australia (BUS) \cite{anaissi2019smart}. 
\end{enumerate}

All the datasets are stored in  a three-way  tensor represented by $sensor \times frequency \times time$. Further details about these datasets are summarized in Table \ref{datasets}. Using these datasets, we run a number of experiment sets to evaluate our proposed FP-CPD method as detailed in the following sections. 

\begin{table}
	\centering
	\caption{Details of datasets}
	\label{datasets}
	\begin{tabular}{l l c r}
		\hline
		Datasets & Size & \begin{tabular}[x]{@{}c@{}}Slice Size\\ $S = \prod_{i=1}^{N-1} I_i$ \end{tabular} & Source \\
		\hline
		BRIDGE & $ \mathcal{X} \in \Re^{24 \times 1200 \times 262}$ & 28,800 & \cite{anaissi2018tensor} \\
		BUILDING &$ \mathcal{X} \in \Re^{24 \times 8192 \times 240}$ & 196,608 & \cite{larson1987alamos} \\
		ROAD &$ \mathcal{X} \in \Re^{96 \times 4666 \times 1826}$ & 447,936 & \cite{schimbinschi2015traffic} \\
		BUS &$ \mathcal{X} \in \Re^{2 \times 2000 \times 5346}$ & 4,000 & \cite{anaissi2019smart} \\
		\hline
	\end{tabular}
\end{table}

\subsection{Evaluating Performance of FP-CPD }
\label{sec:speed}

\begin{figure*}[!t]
	\centering
	\captionsetup[subfloat]{justification=centering}
	\subfloat[BUILDING] 
	{{\includegraphics[height=2.5in,width=3in]{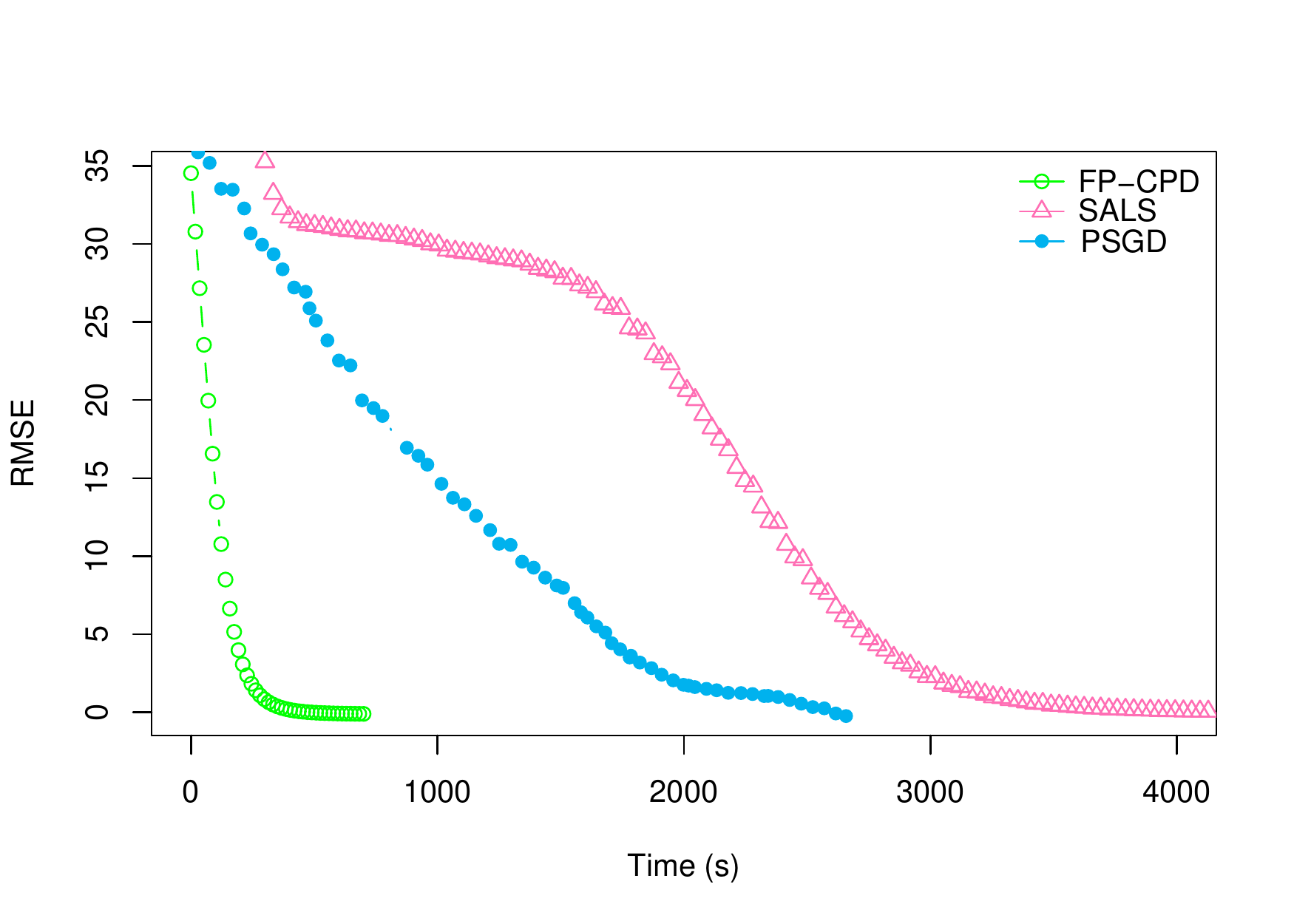} }}	
	\subfloat[BUS] 
	{{\includegraphics[height=2.5in,width=3in]{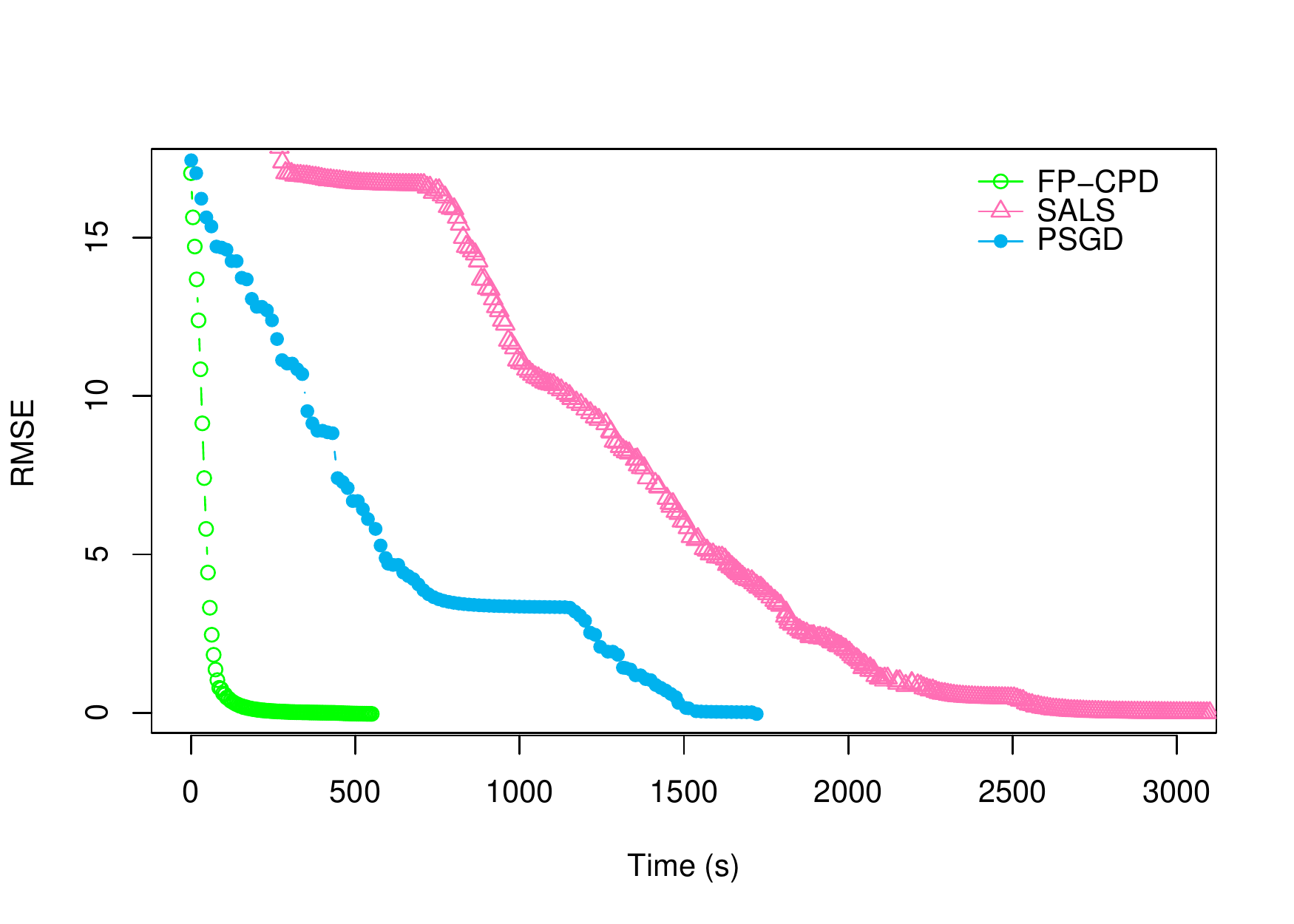} }} \\
	\subfloat[ROAD] 
	{{\includegraphics[height=2.5in,width=3in]{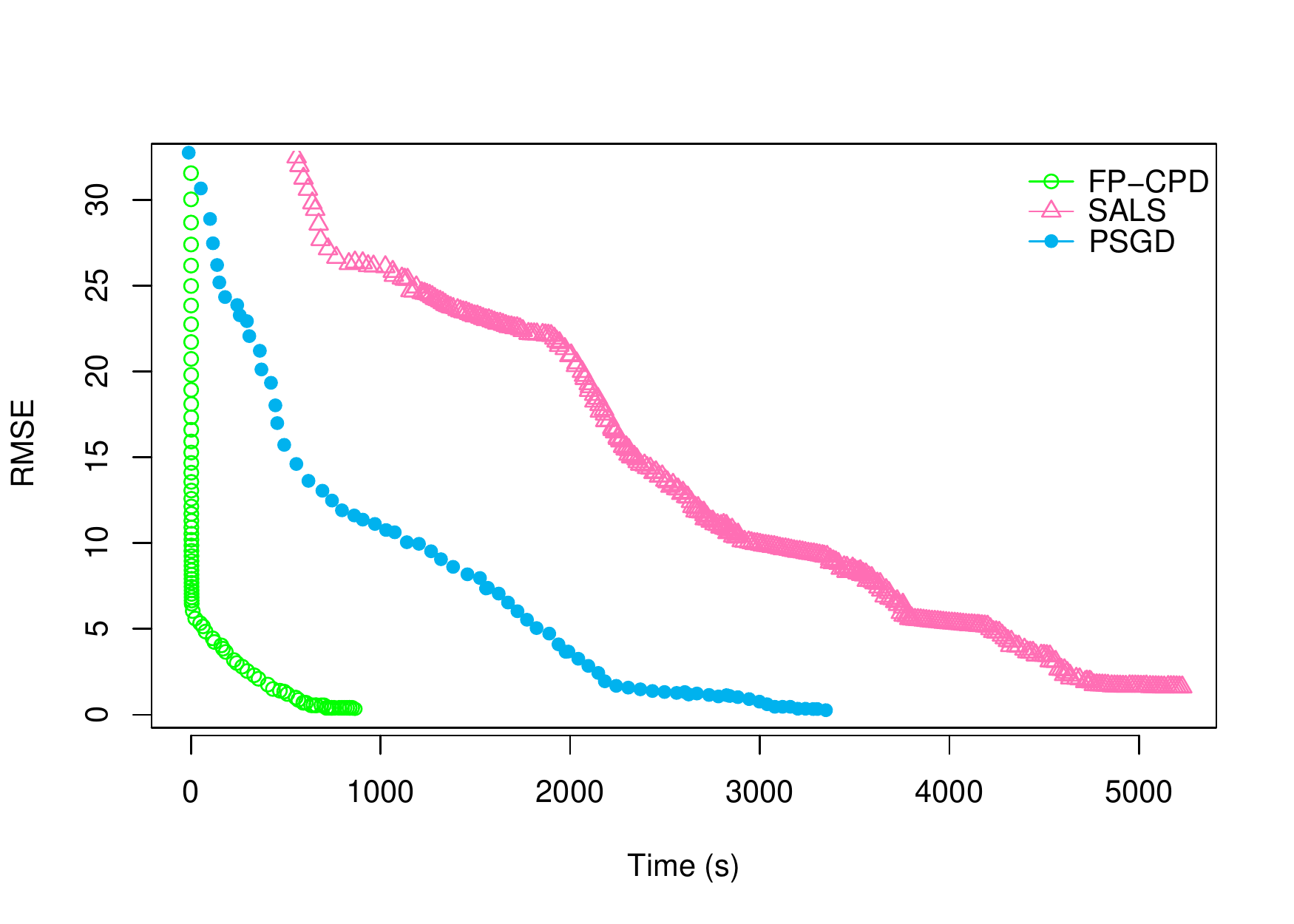} }}%
	\subfloat[ BRIDGE] 
	{{\includegraphics[height=2.5in,width=3in]{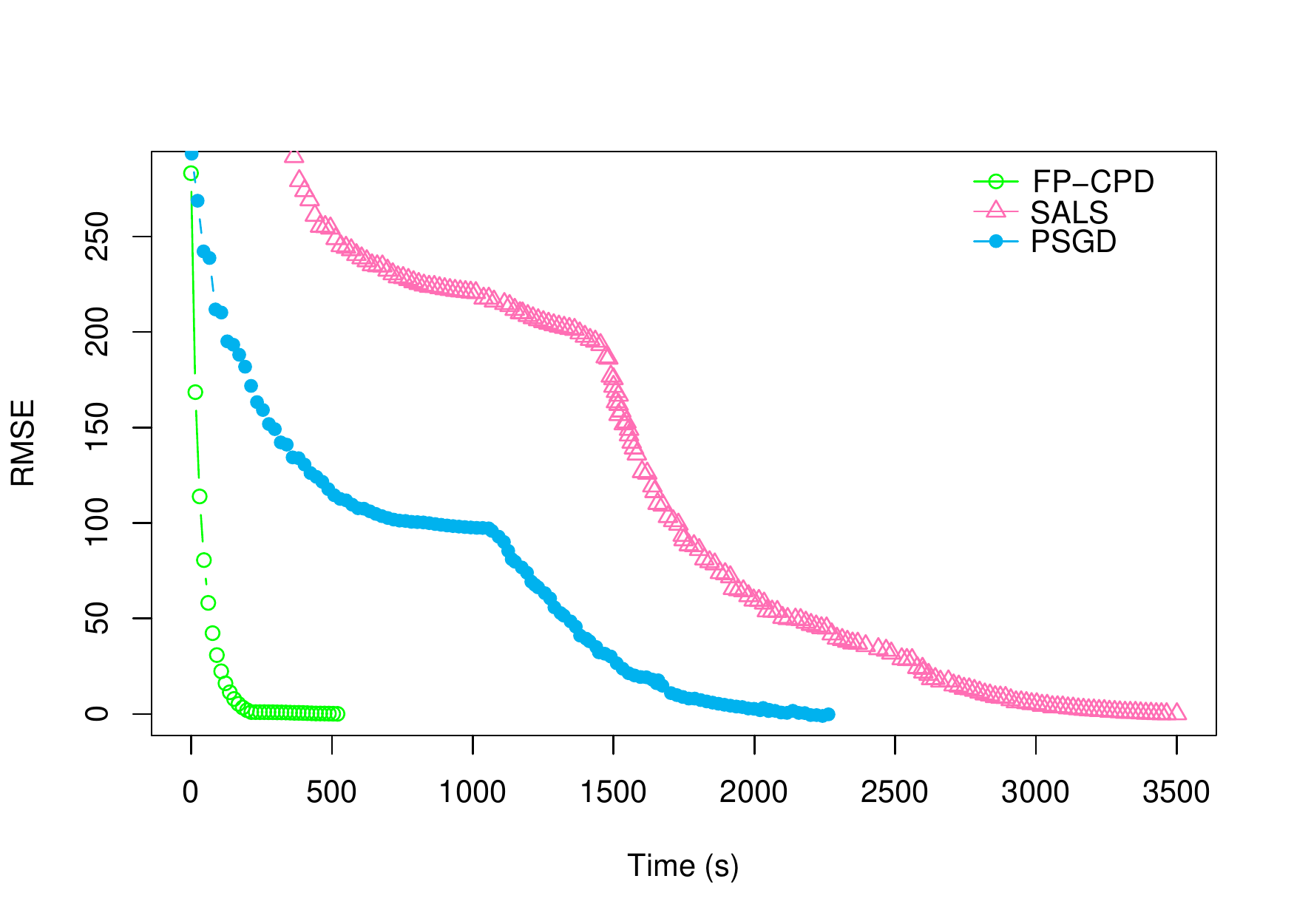} }}%
	\caption{Comprison of training time and RSME of FP-CPD, SALS and PSGD on the four datasets.}%
	\label{comp}
\end{figure*}

The goal of first experiment set is to evaluate the performance of our FP-CPD method in terms of training time error rate. To achieve this, we compare the performance of our proposed FP-CPD and PSGD and SALS algorithms. To make a fair and objective comparison, we implemented the three algorithms under the same experimental settings as described in ~\ref{sec:setup}. We evaluated the performance of each method by plotting the time needed to complete the training process versus the root mean square error (RMSE). We run the same experiment on the four datasets (BRIDGE, BUILDING, ROAD and BUS). Figure \ref{comp} shows the RMSE and the training time of the three algorithms resulted from our experiments. As illustrated in the figure, our FP-CPD algorithm significantly outperformed the PSGD and SALS algorithms in terms of convergence and training speed. The SALS algorithm was the slowest among the three algorithms due to the fact that CP decomposition is a non-convex problem which can be better handled using scholastic methods. Furthermore, another important factor that contributed to the significant performance improvements in our FP-CPD method is the utilization of the Nesterov method along with the perturbation approach in our FP-CPD method. From the first experiment set, it can be concluded that our FP-CPD method is more effective in terms of RMSE and can carry on training faster compared to similar parallel tensor decomposition methods.    

\subsection{Evaluating Effectiveness of FP-CPD} 

\begin{figure*}[!t]
	\centering
	\captionsetup[subfloat]{justification=centering}
	\subfloat[FP-CPD.] {{\includegraphics[height=2.5in,width=3in]{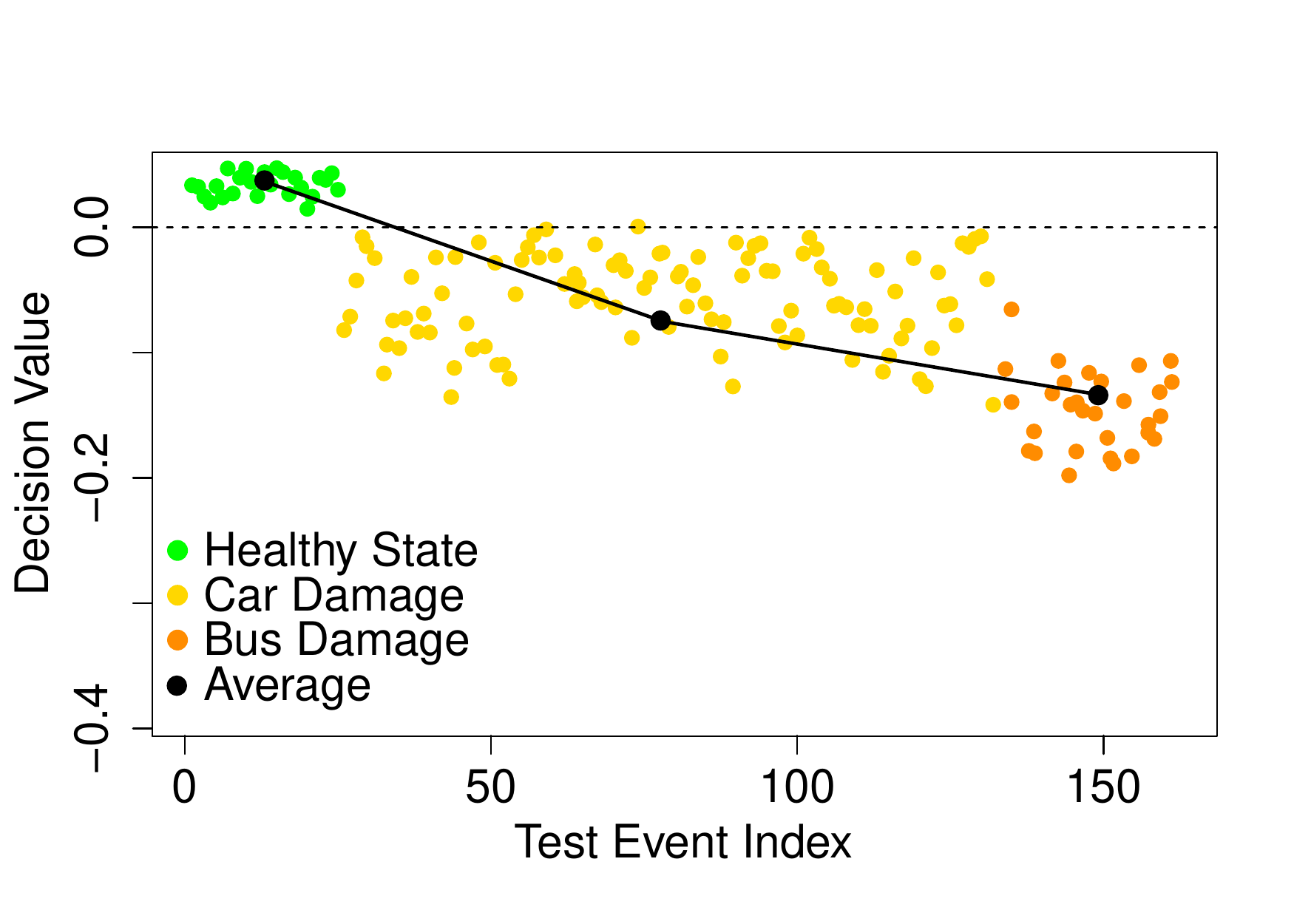} }}	
	\subfloat[ SALS.] 
	{{\includegraphics[height=2.5in,width=3in]{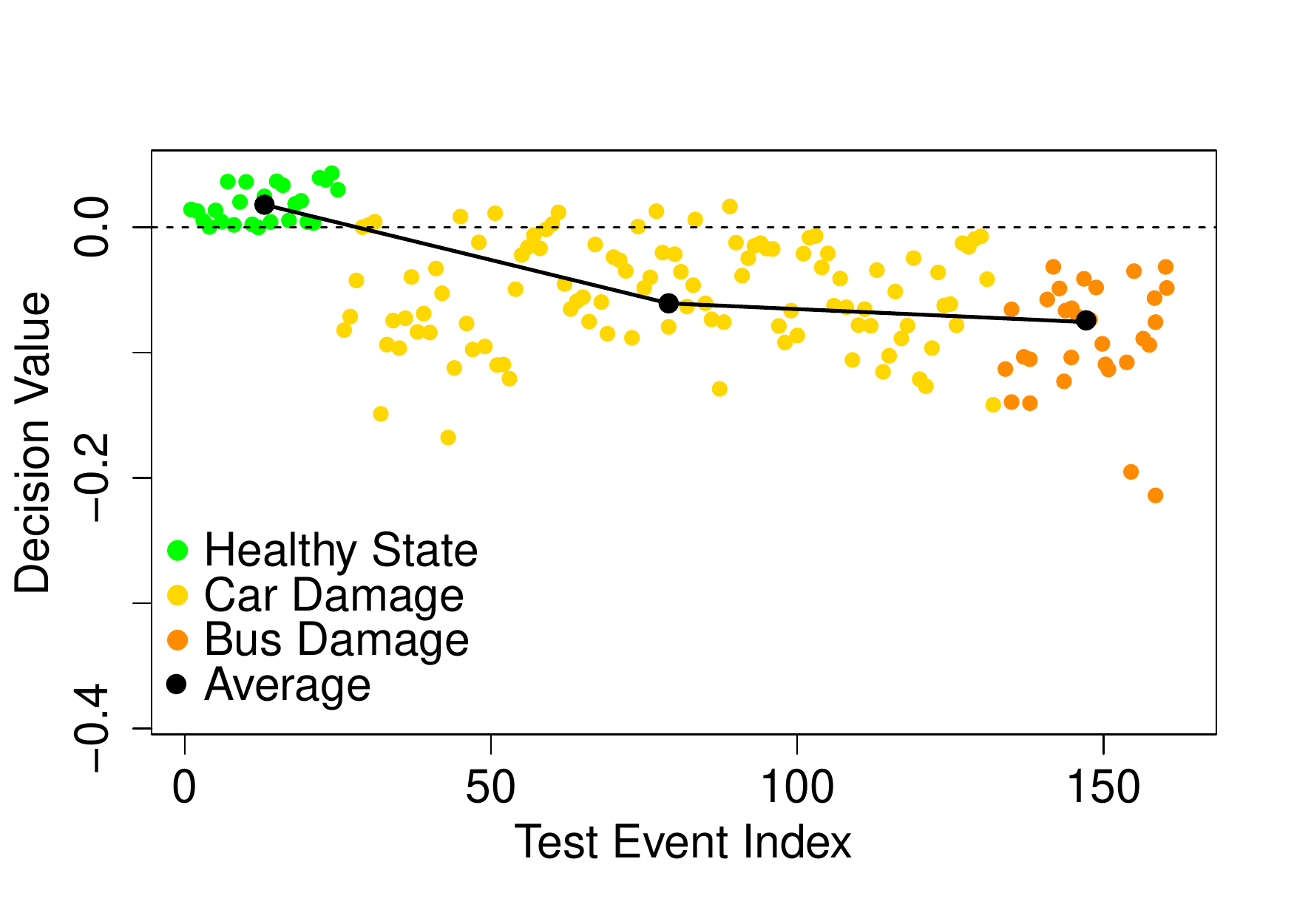} }}%
	\caption{Damage estimation applied on Bridge data using decision values obtained by  one-class SVM.}%
	\label{dv_est}
\end{figure*}
\begin{figure*}[!t]
	\captionsetup[subfloat]{justification=centering}
	\subfloat[FP-CPD.] {{\includegraphics[trim=0cm 1cm 3cm 0cm,clip=true,height=2.5in,width=3.5in]{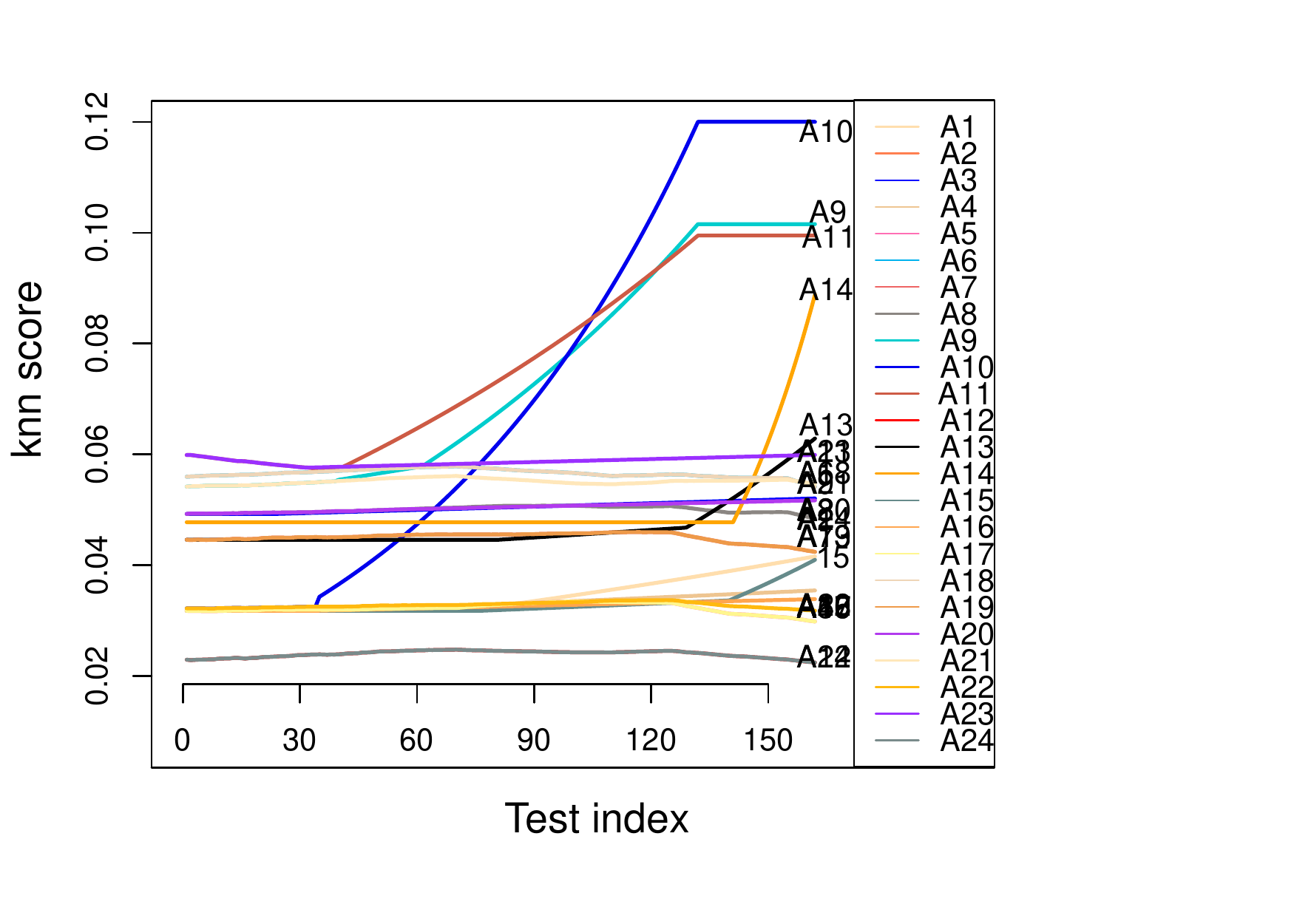} }}	
	\subfloat[ SALS.] {{\includegraphics[trim=0cm 1cm 3cm 0cm,clip=true,height=2.5in,width=3.5in]{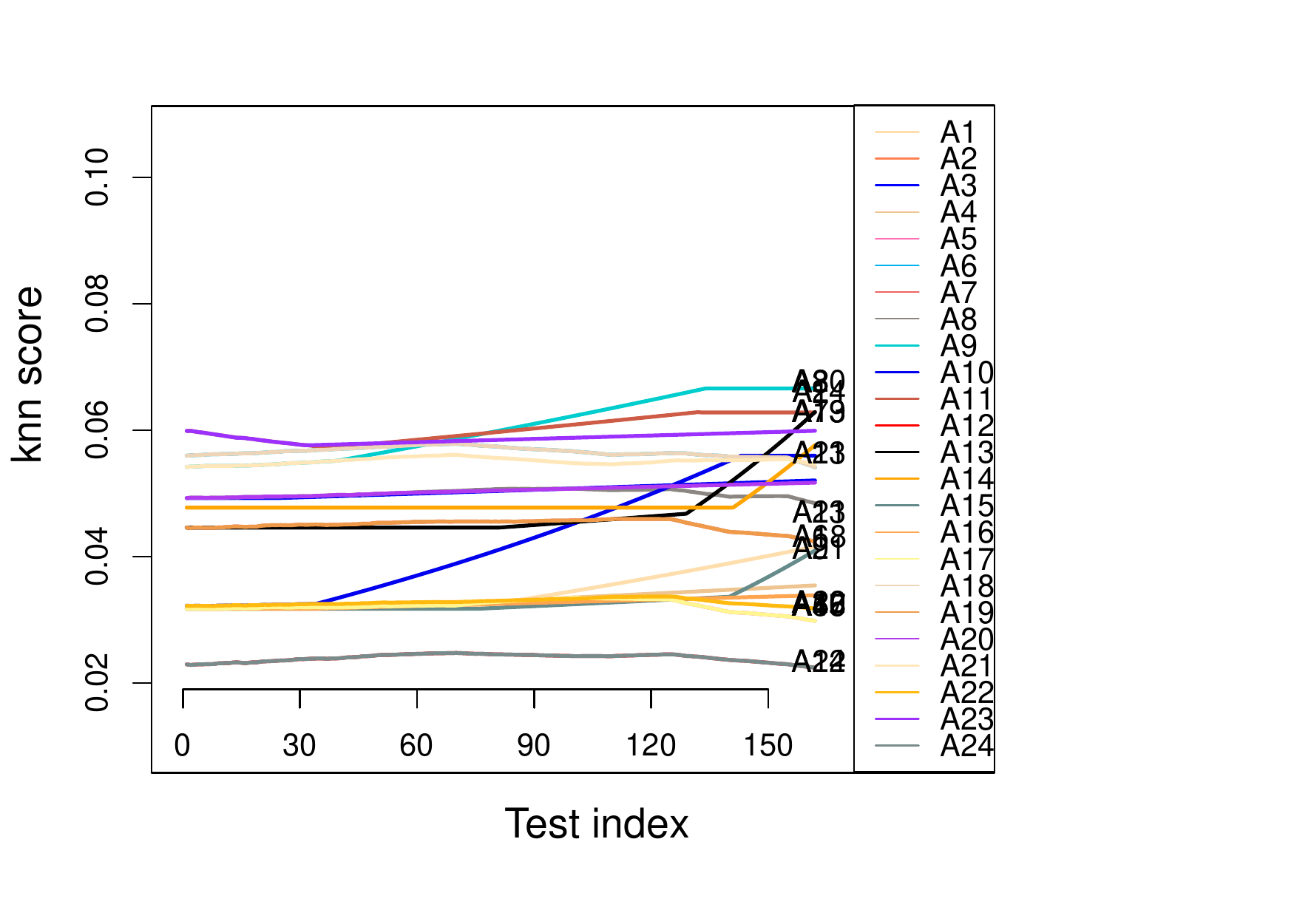} }}%
	
	\caption{ Damage localization for the Bridge data: FP-CPD  successfully localized damage locations.}
	\label{fig:wsu_necpd_loc}
\end{figure*}

\begin{figure*}[!t]
	\centering
	\captionsetup[subfloat]{justification=centering}
	\subfloat[FP-CPD.] {{\includegraphics[height=2.4in,width=2.8in]{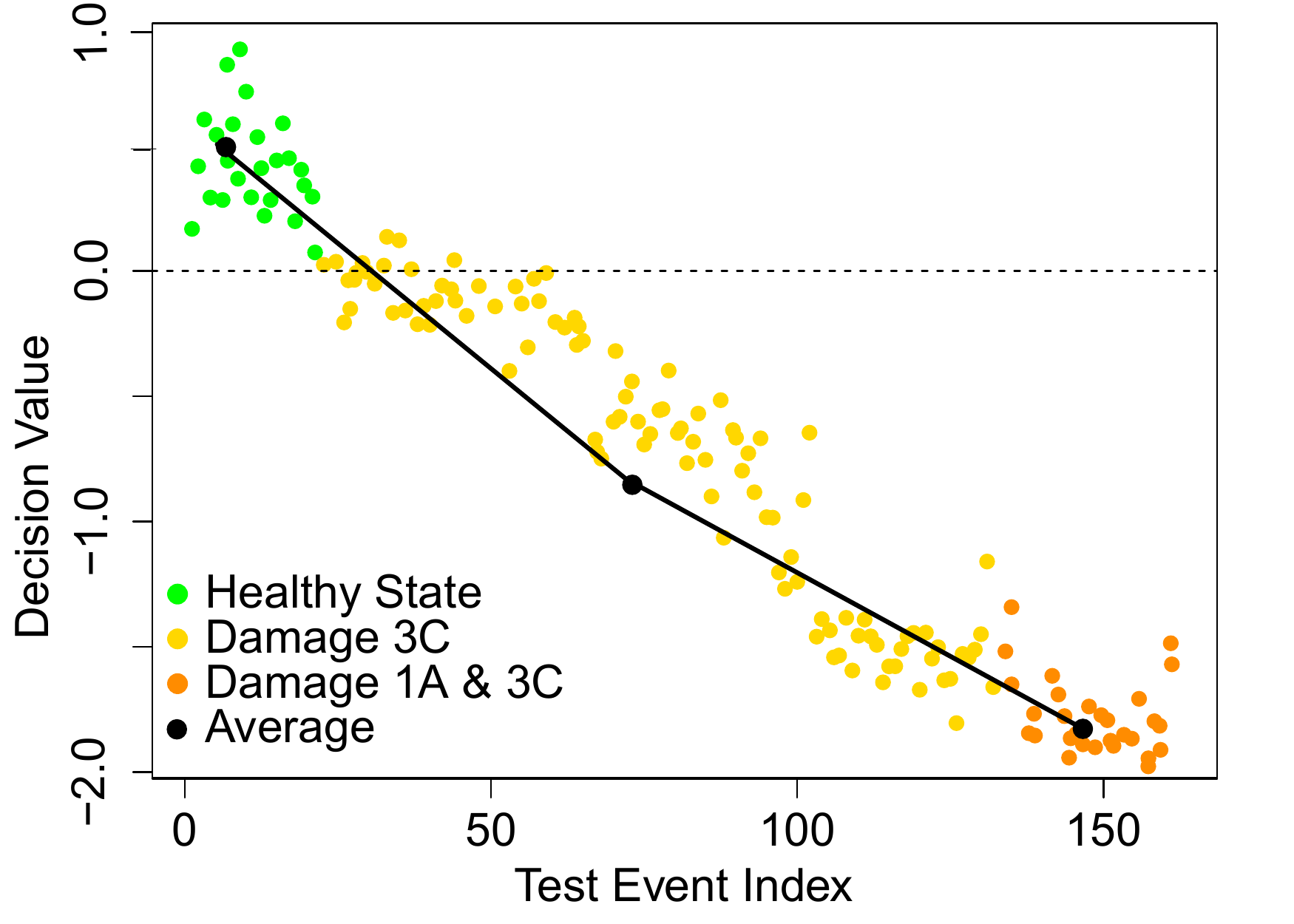} }}
	\subfloat[ SALS.] {{\includegraphics[height=2.4in,width=2.8in]{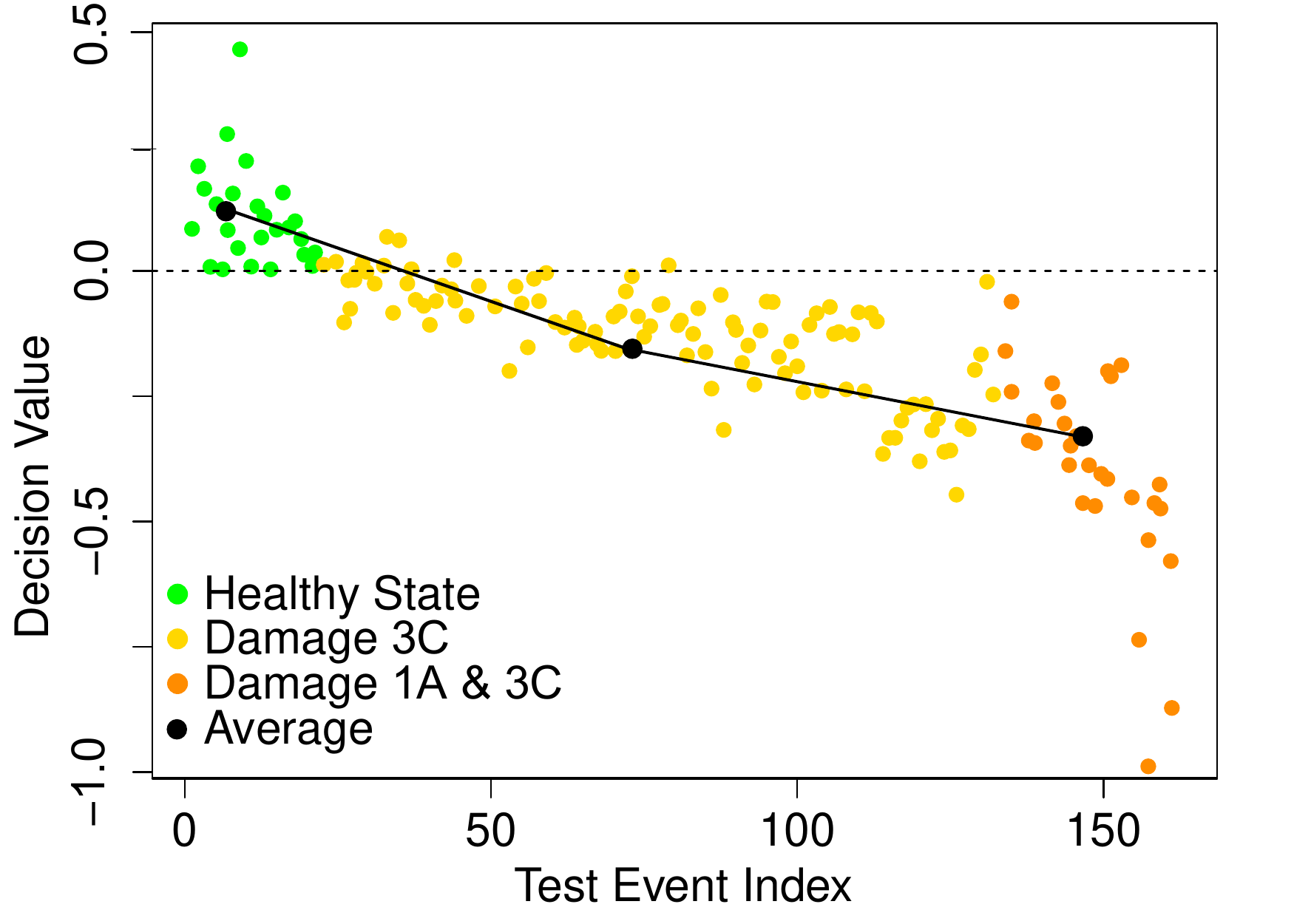} }}%
	\caption{Damage estimation applied on Building data using decision values obtained by  one-class SVM.}%
	\label{fig:dv_build}
\end{figure*}

\begin{figure*}[!t]
	\captionsetup[subfloat]{justification=centering}
	\subfloat[FP-CPD.] {{\includegraphics[height=2.5in,width=3.5in]{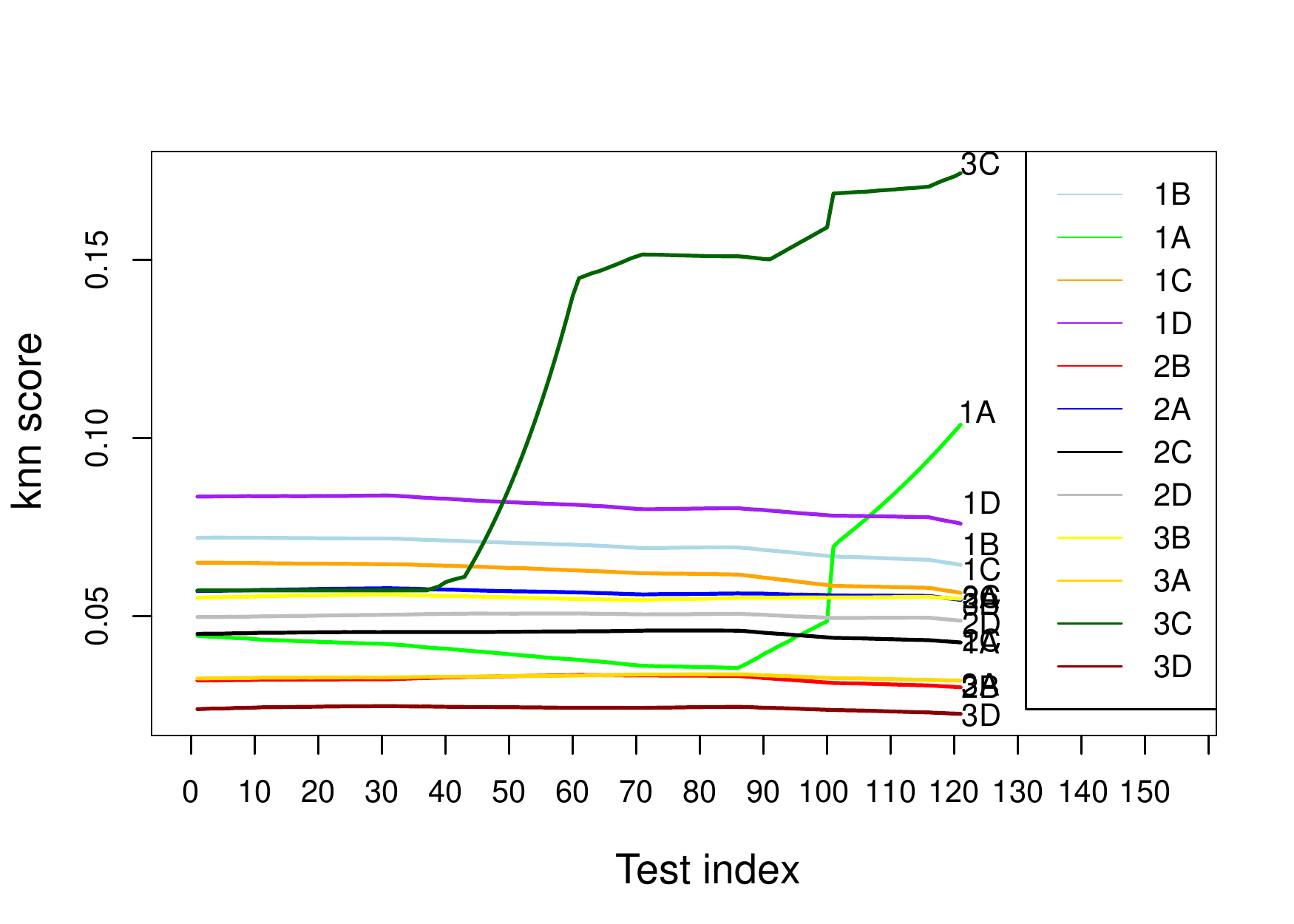} }}	
	\subfloat[ SALS.] {{\includegraphics[height=2.5in,width=3.5in]{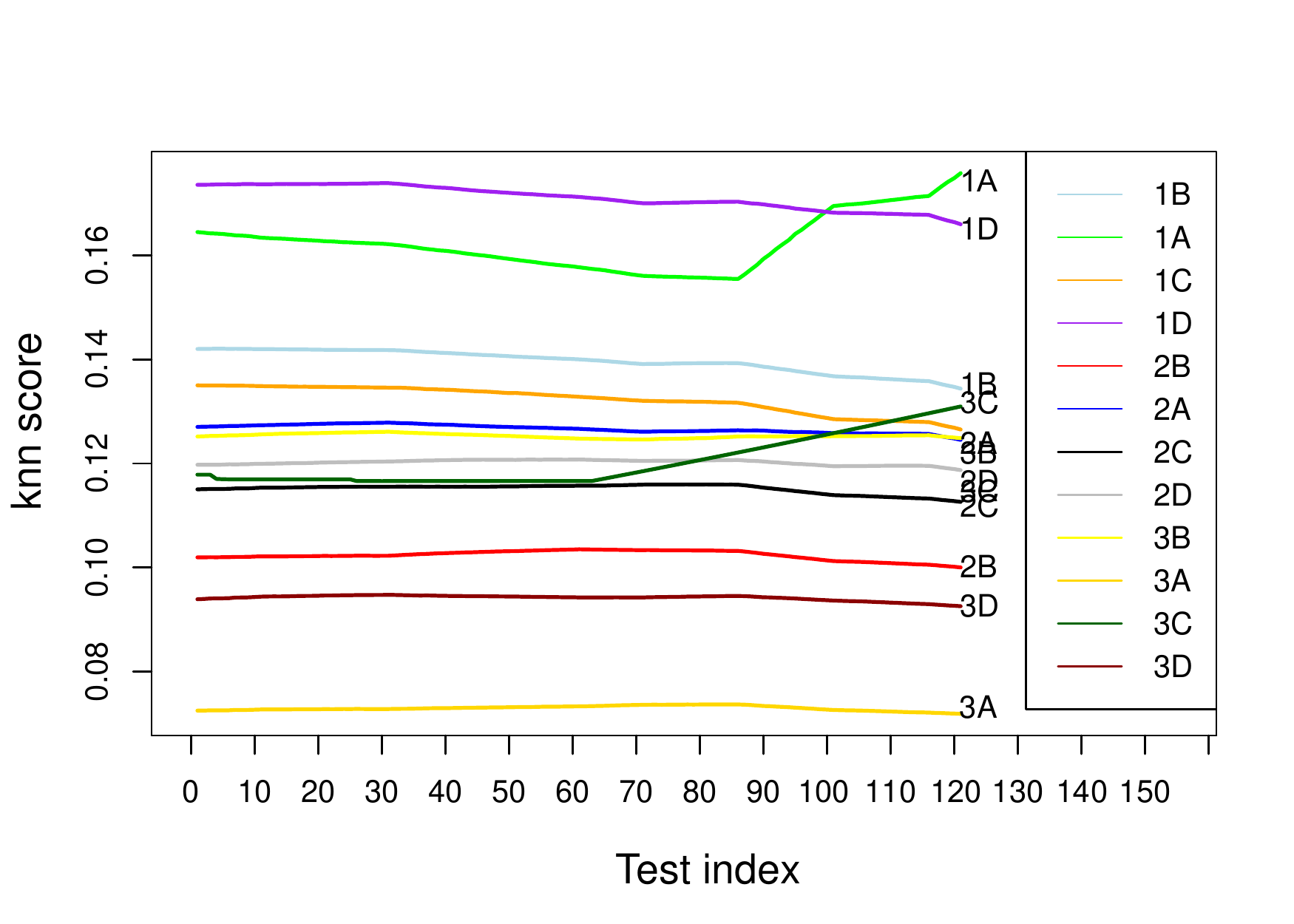} }}%
	
	%\subfloat[ batchCP-ALS.] {{\includegraphics[height=2.2in,width=3.5in]{build_batch_loc} }}%
	\caption{ Damage localization for the Building data: FP-CPD  successfully localized  damage locations.}
	\label{fig:build_necpd_loc}
\end{figure*}

Our FP-CPD method demonstrated better speed and RSME in comparison to PSGD and SALS methods. However, it is still crucial to ensure that the proposed method is also capable of achieving accurate results in practical tensor decomposition problems. Therefore, the second experiment set aims to demonstrate the accuracy of our model in practice, specifically building structures in smart cities. To achieve this, we evaluate the performance of our FP-CPD in terms of its accuracy to detect damage in build and bridge structures, assessing the severity of detected damage and the localization of the detected damage. We carry on the evaluation on the BRIDGE and BUILDING datasets which are explained in the following sections. For comparative analysis, we choose SALS method as a baseline competitor to our FP-CPD. This is because PSGD has similar convergence as FP-CPD but the later takes less time to train as illustrated in section~\ref{sec:speed}. 

\subsubsection{The Cable-Stayed Bridge Dataset}
\label{s:data_wsu}

\begin{figure*}
	\centering
	\includegraphics[scale=0.75]{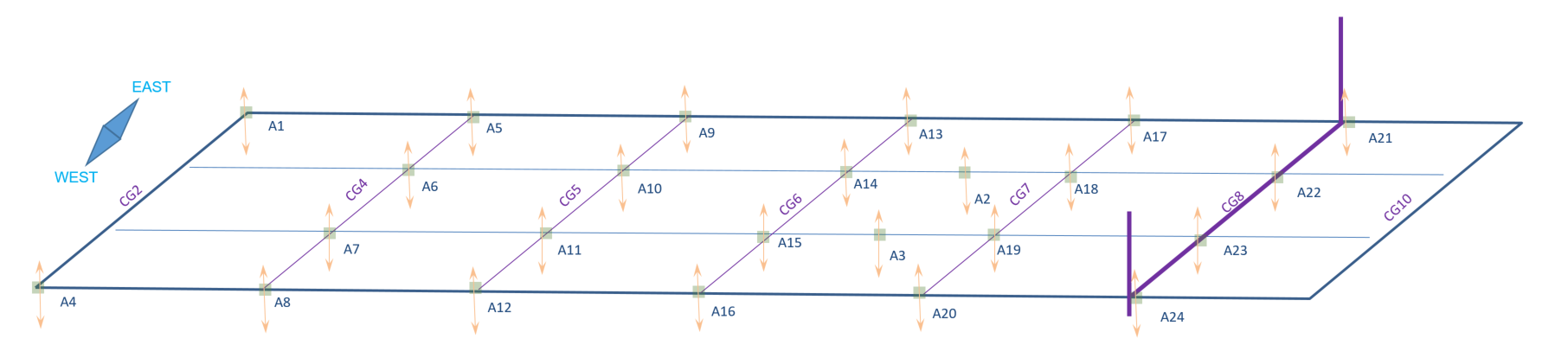}
	\caption{The locations on the bridge's deck of the 24 $Ai$ accelerometers used in the BRIDGE dataset. The cross girder $j$ of the bridge is displayed as $CGj$ \cite{anaissi2018tensor}.}
	\label{fig:wsuloc}
\end{figure*}

In this dataset, 24 uni-axial accelerometers and 28 strain gauges were attached at different locations of the Cable-Stayed bridge to measure the vibration and strain responses of the bridge. Figure~\ref{fig:wsuloc} illustrates the positioning of the 24 sensors on the bridge deck. The data of interest in our study is the accelerations data which were collected from sensors $Ai$ with $i\in [1;24]$. The bridge is in healthy condition. In order to evaluate the performance of damage detection methods, two different stationary vehicles (a car and a bus) with different masses were placed on the bridge to emulate two different levels of damage severity  \cite{kody2013identification,cerda2012indirect}.  The three different categories of data were collected  in that study are: "\textit{Healthy-Data}" when the bridge is free of vehicles;  "\textit{Car-Damage}"  when a  light car vehicle is placed on the bridge close to location $A10$; and  "\textit{Bus-Damage}" when a heavy bus vehicle is located on the bridge at location $A14$. This experiment generates 262 samples (i.e., events) separated into three categories: "\textit{Healthy-Data}" (125 samples), "\textit{Car-Damage}" data (107 samples) and "\textit{Bus-Damage}" data(30 samples). Each event consists of acceleration data for a period of 2 seconds sampled at a  rate of 600Hz. The resultant event's feature vector composed of 1200 frequency values. Figure~\ref{fig:wsuloc} illustrates the setup of the sensors on the bridge under evaluation. 

\subsubsection{The LANL Building Dataset}
\label{s:data_b}
\begin{figure*}
	\centering
	\includegraphics[scale=0.7]{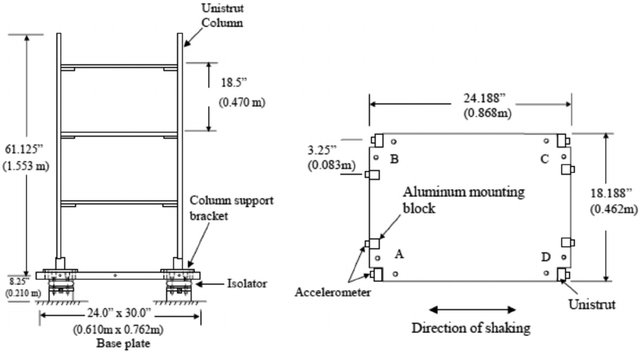}
	\caption{Three-story building and floor layout \cite{larson1987alamos}.}
	\label{fig:alamos}
\end{figure*}

This data is based on experiments conducted by LANL \cite{larson1987alamos} using a specimen for a three-story building structure as shown in Figure \ref{fig:alamos}. Each joint in the building was instrumented by two accelerometers.  The excitation data was generated using a shaker placed at corner $D$. Similarly, for the sake of damage detection evaluation, the damage was simulated by detaching or loosening the bolts at the joints to induce the aluminum floor plate moving freely relative to the Unistrut column. Three different categories of data were collected in this experiment: "\textit{Healthy-Data}" when all the bolts were firmly tightened;  "\textit{Damage-3C}" data when the bolt at location 3C was loosened; and  "\textit{Damage-1A3C}" data when the bolts at locations 1A and 3C were loosened simultaneously. This experiment generates 240 samples (i.e., events) which also were separated into three categories: \textit{Healthy-Data} (150 samples), "\textit{Damage-3C}" data (60 samples) and "\textit{Damage-1A3C}" data(30 samples). The acceleration data was sampled at 1600 Hz. Each event was measured for a period of 5.12 seconds resulting in a vector of 8192 frequency values.

\subsubsection{Feature Extraction}
\label{section:fe}
The raw signals of the sensing data collected in the aforementioned experiments exist in the time domain. In practice, time domain-based features may not capture the physical meaning of the physical structure. Thus, it is important to convert the generated data to a frequency domain. For all the datasets,  we initially normalized the time-domain features to have zero mean and one standard deviation. Then we used the fast Fourier transform method to convert them into the frequency domain. The resultant three-way data collected from the Cable-Stayed Bridge now has a structure of  600 features  $\times$ 24 sensors  $\times$ 262 events. For the LANAL BUILDING dataset,  we computed the difference between signals of two adjacent sensors which resulted in 12 different joints in the three stories as in \cite{larson1987alamos}.  Then we selected the first 150 frequencies as a feature vector which resulted in a three-way data with a structure of 768 features $\times$ 12 locations  $\times$ 240 events.

\subsubsection{Experiments}
\label{section:setup}
For both BUILDING and BRIDGE datasets, we applied the following procedures:
\begin{itemize}
	
	\item  Using the bootstrap technique, we selected 80\% of the healthy samples randomly for training and the remaining 20\% for testing in addition to the damage samples. We computed the accuracy of our FP-CPD model based on the average results over ten trials of the bootstrap experiment.
	
	\item We used the core consistency diagnostic (CORCONDIA) technique described in \cite{bro2003new} to determine the number of rank-one tensors $\mathcal{X}$ in the FP-CPD.
	
	\item   We used the one-class support vector machine (OSVM) \cite{scholkopf2000support} as a model for anomaly detection. The Gaussian kernel parameter  $\sigma$ in OCSVM is  tuned using the  Edged Support Vector (ESV) algorithm \cite{anaissi2018gaussian}, and the rate of anomalies $\nu$  was set to 0.05. 
	
	\item  We used the $\Fscore$  measure to compute the accuracy of data values resulted from our model for damage detection. It is defined as $\textrm{\Fscore} = 2 \cdot \dfrac{\textrm{Precision}  \times \textrm{Recall} }{\textrm{Precision} + \textrm{Recall}}$ where $\textrm{Precision} = \dfrac{\textrm{TP} }{\textrm{TP} + \textrm{FP}}$ and $\textrm{Recall}  = \dfrac{\textrm{TP} }{\textrm{TP} + \textrm{FN}}$ (the number of true positive, false positive and false negative are abbreviated by TP, FP and FN, respectively). 
	
	\item We compared the results of the competitive method SALS proposed in \cite{maehara2016expected}  against the ones resulted from our FP-CPD method.
	
\end{itemize}

\subsubsection{Results and Discussion}
\label{sec:results}

\subsubsection{The Cable-Stayed Bridge Dataset:}
\label{s:wsueval}
Our FP-CPD method with one-class SVM was initially validated using the vibration data collected from the cable-stayed bridge (described in Section \ref{s:data_wsu}). The healthy training three-way tensor data (i.e., \textbf{training} set) was in the form of $ \mathcal{X} \in \Re^{24 \times 600 \times 100}$. The 137 examples related to the two damage cases were added to the remaining 20\% of the healthy data to form a \textbf{testing} set, which was later used for model evaluation. We conducted the experiments as followed the steps described in Section\label{section:setup}. As a result, this experiment generates a damage detection  accuracy $\Fscore$  of $1 \pm 0.00$ on the \textbf{testing} data. On the other hand, the $\Fscore$     accuracy of one-class SVM using SALS  is recorded at $0.98 \pm 0.02$.

As demonstrated from the results of this experiment, the tensor analysis with our proposed FP-CPD is capable to capture the underlying structure in multi-way data with better convergence. This is further illustrated by plotting the decision values returned from one-class SVM based FP-CPD (as shown in Figure \ref{dv_est}). We can clearly separate the two damage cases ("Car-Damage" and "Bus-Damage") in this dataset where the decision values are further decreased for the samples related to the more severe damage cases (i.e., "Bus-Damage"). These results suggest using the decision values obtained by our FP-CPD and one-class SVM as structural health scores to identify the damage severity in a one-class aspect. In contrast, the resultant decision values of one-class SVM based on SALS  are also able to track the progress of the damage severity in the structure but with a slight decreasing trend in decision values for "Bus-Damage" as shown in Figure \ref{dv_est}.

The last step in this experiment is to analyze the location matrix $B$ obtained from FP-CPD to locate the detected damage. Each row in this matrix captures meaningful information for each sensor location. Therefore,  we calculate the average distance from each row in the matrix $B_{new}$ to $k$-nearest neighboring rows. Figure \ref{fig:wsu_necpd_loc} shows the obtained $k$-nn score for each sensor. The first 25 events (depicted on the x-axis) represent healthy data, followed by 107 events related to "Car-Damage" and 30 events to "Bus-Damage". It can be clearly observed that   FP-CPD method can localize the damage in the structure accurately. Whereas, sensors $A10$ and $A14$ related to the "Car-Damage" and "Bus-Damage" respectively behave significantly different from all the other sensors apart from the position of the introduced damage. In addition, we observed that the adjacent sensors to the damage location (e.g $A9$, $A11$, $A13$ and $A15$) react differently due to the arrival pattern of the damage events. The SALS method, however, is not able to accurately locate the damage since it fails to update the location matrix $B$ incrementally.

\subsubsection{The Building Dataset:}

Following the experimental procedure described in Section \label{section:setup}, our second experiment was conducted using the acceleration data acquired from 24 sensors instrumented on the three-story building  as described in Section \ref{s:data_b}. The healthy three-way data (i.e., \textbf{training} set) is in the form of  $ X \in \Re^{12 \times 768 \times 120}$. The remaining 20\% of the healthy data and the data obtained from the two damage cases were used for testing (i.e., \textbf{testing} set). The experiments we conducted using FP-CPD with one-class SVM have achieved an $\Fscore$ of $95 \pm 0.01$  on the \textbf{testing} data compared to $0.91 \pm 0.00$ obtained from one-class SVM and SALS experiments. 

Similar to the BRIDGE dataset, we further analyzed the resultant decision values which were also able to characterize damage severity. Figure \ref{fig:dv_build} demonstrates that the more severe damage to the $1A$ and $3C$ location test data, the more deviation from the training data with lower decision values. %Check the meaning%    

%that the more severe damage test data related to locations $1A$ and $3C$ were more deviated from the training data with lower  decision values.

Similar to the BRIDGE dataset, the last experiment is to compute the $k$-nn score for each sensor based on the $k$-nearest neighboring of the average distance between each row of the matrix $B_{new}$. Figure \ref{fig:build_necpd_loc} shows the resultant $k$-nn score for each sensor. The first 30 events (depicted on the x-axis) represent the healthy data, followed by 60 events describing when the damage was introduced in location $3C$. The last 30 events represent the damage occurred in both locations $1A$ and $3C$. It can be clearly observed that the FP-CPD  method is capable to accurately localize the structure's damage where sensors $1A$ and $3C$ behave significantly different from all the other sensors apart from the position of the introduced damage. However, the SALS method is not able to locate that damage since it fails to update the location matrix $B$  incrementally.

In summary, the above experiments on the four real datasets demonstrate the effectiveness of our proposed FP-CPD method in terms of time needed to carry out training during tensor decomposition. Specifically, our FP-CPD significantly improves speed of model training and error rate compared to similar parallel tensor decomposition methods, PSGD and SALS. Furthermore, the other experiment sets on the BRIDGE and BUILDING datasets showed empirical evidence of the ability of our model to accurately carry on tensor decomposition on practical case studies. In particular, the experimental results demonstrated that our FP-CPD is able to detect damage in the build and bridge structures, assess the severity of detected damage and localize of the detected damage more accurately than SALS method. Therefore, it can be concluded that our FP-CPD tensor decomposition method is able to achieve faster tensor model training with minimal error rate while carrying on accurate tensor decomposition in practical cases. Such performance and accuracy gains can be beneficial for many parallel tensor decomposition cases in practice especially in real-time detection and identification problems. We demonstrated such benefits with real use cases in structural health monitoring namely building and bridge structures.

\section{Conclusion}
\label{s:conclusion}

%In this paper, we introduce our new method NeCPD for tensor decomposition. The method derived based on well-known CP decomposition with stochastic gradient descent algorithm for multi-way analysis. 

This paper investigated the CP decomposition with a stochastic gradient  descent  algorithm for multi-way data analysis. This leads to a new method named Fast Parallel-CP Decomposition (FP-CPD) for tensor decomposition. The proposed method guarantees the convergence for a given non-convex problem by modeling the second order derivative of the loss function and incorporating little noise to the gradient update. Furthermore, FP-CPD employs Nesterov's method to compensate for the optimization process's delays and accelerate the convergence rate. Based on laboratory and real datasets from the area of SHM, our FP-CPD, with a one-class SVM model for anomaly detection, achieve accurate results in damage detection, localization, and assessment in online and one-class settings. Among the key future work is how to parallelize the tensor decomposition with FP-CPD. Also, it would be useful to apply FP-CPD with datasets from different domains.

%This paper investigated the CP decomposition with stochastic gradient  descent  algorithm for multi-way data analysis. This leads to a new algorithm named NeCPD for tensor decomposition which guarantees  the convergence for a given non-convex problem    by computing  the second order derivative of the loss  function and adding  little noise to the gradient update. It further applies Nesterov's   method   to compensate the delay of the optimization process and  accelerate the  convergence rate. This  new method with one-class SVM model for anomaly detection,  achieved promising results in damage detection, localization, and assessment in an online and one-class manner using laboratory-based and real-life structural  datasets in the area of SHM. Our future works will focus on how to parallelize the tensor decomposition with  NeCPD.

\begin{acknowledgements}
	The authors wish to thank the Roads and Maritime Services (RMS) in New South Wales, New South Wales Government in Australia and  Data61
	(CSIRO) for provision of the support and testing facilities for this research
	work. Thanks are also extended to Western Sydney University for facilitating the experiments
	on the cable-stayed bridge.
\end{acknowledgements}
% Bibliography
\bibliographystyle{plain}
\bibliography{myBib}

\end{document}